\documentclass[11pt, a4paper, logo,
copyright]{deepmind}

\pdftrailerid{redacted}

\makeatletter
\renewcommand\bibentry[1]{\nocite{#1}{\frenchspacing\@nameuse{BR@r@#1\@extra@b@citeb}}}
\makeatother

\usepackage{kantlipsum, lipsum}
\usepackage{dsfont}
\usepackage{dm-colors}
\usepackage{todonotes}
\usepackage{graphicx}
\usepackage{subcaption}
\usepackage{caption}
\usepackage{siunitx}
\usepackage{multirow}

\usepackage[authoryear, sort&compress, round]{natbib}

\DeclareMathOperator*{\argmax}{\arg\!\max}

\graphicspath{{figures/}}

\title{A Generalist Dynamics Model for Control}
\correspondingauthor{ingmar.schubert@tu-berlin.de}

\reportnumber{} % Leave blank if n/a 

\author[*,1]{Ingmar Schubert}
\author[2]{Jingwei Zhang}
\author[2]{Jake Bruce}
\author[2]{Sarah Bechtle}
\author[2]{Emilio Parisotto}
\author[2]{Martin Riedmiller}
\author[2]{Jost Tobias Springenberg}
\author[2]{Arunkumar Byravan}
\author[2]{Leonard Hasenclever}
\author[2]{Nicolas Heess}

\affil[1]{TU Berlin}
\affil[2]{DeepMind}
\affil[*]{Work done at DeepMind}
\begin{abstract}
We investigate the use of transformer sequence models as dynamics models (TDMs) for control.
We find that TDMs exhibit strong generalization capabilities to unseen environments, both in a few-shot setting, where a generalist TDM is fine-tuned with small amounts of data from the target environment, and in a zero-shot setting, where a generalist TDM is applied to an unseen environment without any further training.
Here, we demonstrate that generalizing system dynamics can work much better than generalizing optimal behavior directly as a policy.
Additional results show that TDMs also perform well in a single-environment learning setting when compared to a number of baseline models.
These properties make TDMs a promising ingredient for a foundation model of control.
\end{abstract}
\begin{document}
\maketitle
\section{Introduction}
\begin{figure}[h]
	\centering
	\includegraphics[width=\columnwidth]{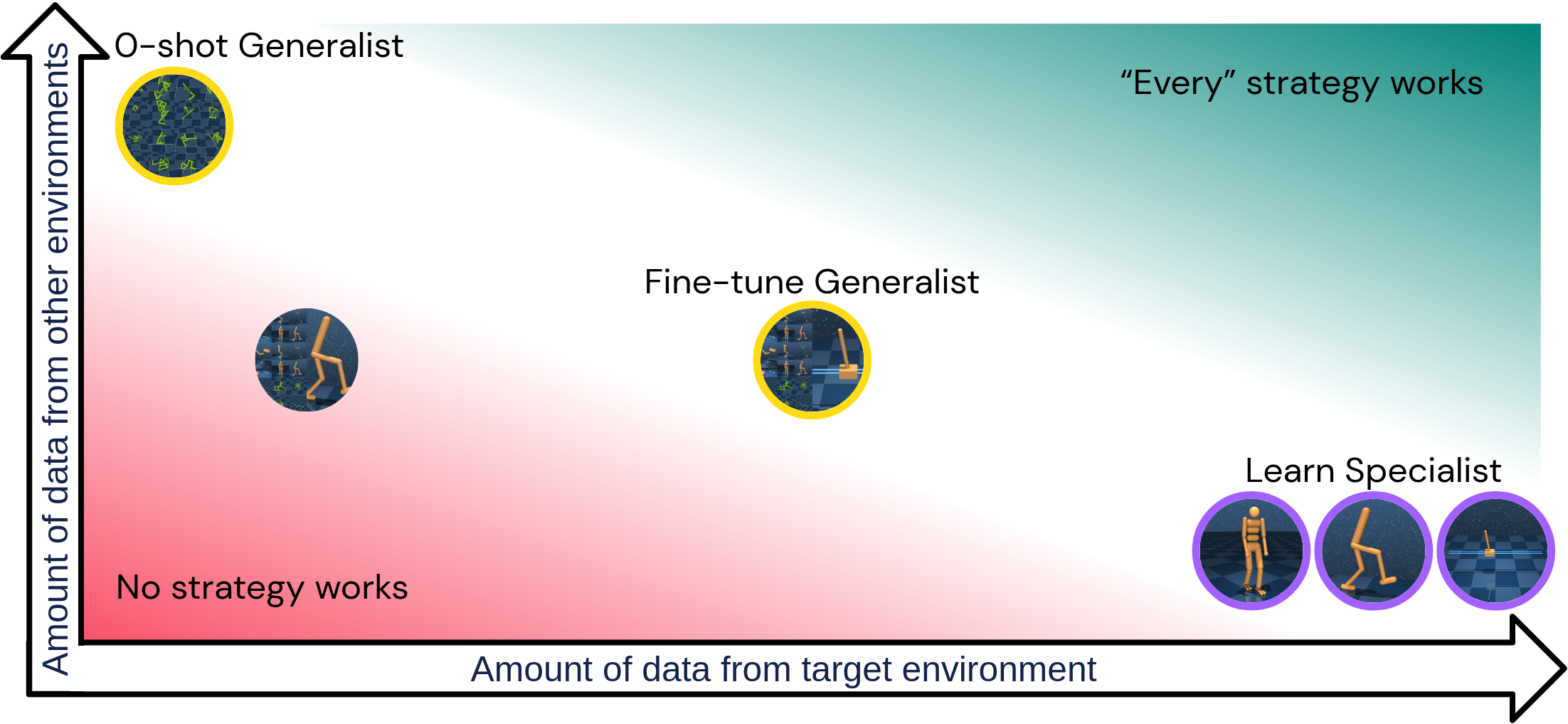}
	\caption{
	Schematic overview of the data regimes for which we show experimental results.
	These regimes are characterized by how much data from the target environment is available to the agent, and how much (potentially generalizable) experience has been collected in other environments.
    The experiments both demonstrate that TDMs are capable single-environment models (marked purple) and generalize across environments (marked yellow).
    If sufficient data from the target environment is available, we can learn a single-environment specialist model (section \ref{sec:expert_model_results}).
    If there are only small amounts of data from the target environment, but more data from other environments, a generalist model can be pre-trained and then fine-tuned on the target environment (section \ref{sec:finetuning_generalist_results}).
    Finally, if we are able to train a generalist model on large amounts of data from different environments, we can zero-shot apply this model to our target environment without fine-tuning (section \ref{sec:zero-shot-generalization-results}).
    We also show an example for unsuccessful generalization (no color) in section \ref{sec:walker_generalization_negative_example}.
	}
	\label{fig:experiments_overview}
\end{figure}
\newpage
An important goal of robotics research is to create embodied agents that are able to achieve a wide range of flexibly defined goals in a wide range of complicated environments.
During the last decade, advancements in artificial intelligence, specifically the renaissance of neural networks, have strongly influenced the field.
Examples include deep visuomotor policies \citep{levine2016end}, dexterous manipulation \citep{andrychowicz2020learning} or multi-agent soccer with humanoid robots \citep{haarnoja2023learning}.
These works have in common that they demonstrate high-quality behavior for complicated tasks, but require large amounts of data, and result in specialist agents.
Broadly speaking, a quality that many state-of-the art approaches to robotics lack is \textit{generality}: the ability to generalize previous experience to unseen environments\footnote{We use ``environment'' in the general sense here: A different environment transition function constitutes a different environment, regardless of whether this difference is due to a change of the robot itself or its surroundings.}.

Recently, training large models on large amounts of data has enabled big leaps in generality in areas such as language modelling \citep{vaswani2017attention,chowdhery2022palm,openai2023gpt4}.
This has inspired interest in using large models to improve generality of embodied agents as well;
either by using language models for high-level decision making (e.g., \citealt{huang2023grounded, driess2023palme}) or by using the large model itself to output control instructions (e.g., \citealt{reed2022generalist}).

The present work focuses on the latter approach - using large models, specifically transformer sequence models, for control.
While most previous work considers using transformers for policy learning, we study their use as dynamics models, an approach we refer to as transformer dynamics models (TDMs).
Traditionally, the motivation for learning explicit dynamics models and using them for control is that the dynamics are independent of the goal. 
Therefore, once learned, a dynamics model can be reused for creating optimal behavior with respect to multiple goals.
In this work, we demonstrate an additional advantage: In certain situations, a dynamics model generalizes better than a behavior policy to unseen environments (not only unseen goals), thus enabling us to create model-based generalist agents that generalize better than their model-free counterparts.

Concretely, we highlight two different aspects of TDMs in our experiments (see overview in Fig.\ \ref{fig:experiments_overview}):
First, we demonstrate that TDMs generalize strongly across environments; specifically, we show that a generalist TDM can be used for few-shot or even zero-shot generalization to unseen environments.
Second, we demonstrate that, compared to a number of baselines, TDMs make accurate predictions suitable for planning when learning from transition data of the target environment (specialist model learning).
Our contributions are as follows:
\begin{enumerate}
    \item We use transformer sequence models as TDMs for control, and we describe a simple setup to evaluate learned models in an MPC loop together with a random shooting planner.
    \item Our main results are in the generalist setting, i.e., when training the TDM on transition data from environments different from the target environment. Here we find strong generalization capabilities, both few-shot and zero-shot:
    \begin{enumerate}
        \item In a few-shot setting (fine-tuning a generalist), we observe strong generalization effects, which can be exploited to obtain a good dynamics model given limited data.
        In our experiments, this approach surpasses even lightweight specialist models (see section \ref{sec:finetuning_generalist_results}).
        \item In a zero-shot setting, we observe that the generalist TDM generalizes substantially better than its generalist policy counterpart (see section \ref{sec:zero-shot-generalization-results}).
    \end{enumerate}
    \item While not our main focus, we also investigate TDMs in the specialist setting, i.e., when trained on transition data from the target environment.
    Here we observe that TDMs make accurate predictions suitable for planning in a range of difficult control tasks, and outperform a number of baseline models (section \ref{sec:expert_model_results}).
\end{enumerate}

\section{Related work}
\label{sec:related_work}
\subsection{Learned models for decision making and model-based reinforcement learning}
Model-based decision making algorithms \citep{moerland2023model} use in their decision making an explicit (often learned) dynamics model of the environment they operate in.
We can distinguish between planning approaches that then use this model to obtain local solutions of optimal behavior and model-based reinforcement learning (RL) approaches that obtain global solutions (or policies).
Examples of the former category include \citet{watter2015embed, schrittwieser2020mastering, chua2018deep, lutter2021learning, zhang2023leveraging, park2023predictable}, and we compare with PETS \citep{chua2018deep} in our experiments.
Examples of the latter are \citet{ha2018recurrent, heess2015learning, kaiser2019model, gelada2019deepmdp, hafner2019dream, hafner2020mastering, byravan2021evaluating, yin2022planning}, and we compare with the dynamics model of Dreamer V2 \citep{hafner2020mastering} in our experiments.
In both cases, we observe better results for TDMs.

\subsection{Transformers for decision making}
The idea to use transformer sequence models for decision making in sequential decision problems has gained a lot of traction lately.
\citet{parisotto2020stabilizing} introduce architecture features that allow for stable training of transformers with RL objectives.
% Decision Transformer
The Decision Transformer \citep{chen2021decision} is trained to model the joint distribution of observations, actions, and returns, and generates high-return behavior by conditioning on high returns.
% Trajectory transformer
The Trajectory Transformer \citep{janner2021offline} is trained in a similar way, and is then conditioned in different ways for imitation learning, goal-conditioned reinforcement learning (RL) and offline RL.
% TAP
\citet{jiang2022efficient} address unfavorable scaling of \citet{janner2021offline} to high dimensions by introducing a learned latent space.
In \citet{micheli2022transformers,robine2023transformer}, a transformer is used to learn a world model \citep{ha2018world}, which is then used to train a policy using RL inside it.
This is similar in spirit to our approach, in that we also explicitly use TDMs to predict the system's dynamics.
However, to our knowledge, the present work is the first to investigate generalization of TDMs across environments.
Other distinctions are that we use TDMs not to train a global RL agent, but for local decision making with MPC, and that we focus on control problems typical for robotics, rather than Atari \citep{bellemare2013arcade}.

\subsection{General control agents}
General control agents are agents that are able to successfully operate in different environments.
System identification for control \citep{aastrom1971problems,van1995identification,ljung1999system} can be seen as early approaches to such generalist agents.
A more recent line of works represents generalist agents as graph neural networks \citep{scarselli2008graph, battaglia2018relational}.
Examples of this are \citet{wang2018nervenet, huang2020one, blake2021snowflake, sanchez2018graph}.
Most recently, there has been increased interest in using transformers \citep{kurin2020my, gupta2022metamorph, brohan2022rt, sun2023smart, yang2023foundation, furuta2022system}.
Gato \citep{reed2022generalist} is a generalist sequence model that, in addition to being used as a generalist control policy for a wide variety of control problems, can also perform many other tasks like image captioning and acting as a chat bot.
Our work is based on the Gato architecture, and in this sense it is most closely related to this work.
However, in all control tasks in \citet{reed2022generalist}, the model is used as a behavior cloning (BC) policy.
In the present work, we use the model as a dynamics model for planning in an MPC loop.
While learning models and policies from trajectory data is not mutually exclusive, and combining both can make sense (section \ref{sec:humanoid_proposal}), we also demonstrate that, at least for some problems, TDMs can generalize significantly better than policies (section \ref{sec:zero-shot-generalization-results}).

\section{Background}
\label{sec:backgroud}
\subsection{Modelling trajectory data with transformers}
\label{sec:background_modelling_trajectory}
This work is based on the Gato transformer architecture first published in \citet{reed2022generalist}.
The transformer model with parameters $\theta$ models the joint distribution of a sequence of integer tokens $(T_1,...,T_q)$ autoregressively as
\begin{align}
    p_\theta(T_1,...,T_q) = \Pi_{i=1}^q p_\theta(T_i|T_1,...,T_{i-1}) \quad \text{.}
\end{align}
The model is fitted to the conditional distribution $p(T_i|T_1,...,T_{i-1})$ by minimizing the negative log-likelihood loss
\begin{align}
    \mathcal{L}(\theta) = -\sum_{i=1}^q \log p_\theta(t_i|t_1, ..., t_{i-1}) \quad \text{,}
\end{align}
where $(t_1, ..., t_q) \sim (T_1,...,T_q)$ is a sequence of tokens from the data.

\begin{figure}[h]
	\centering
	\includegraphics[width=\columnwidth]{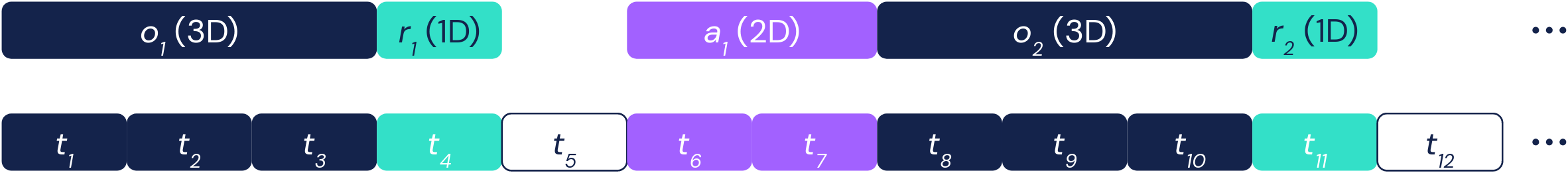}
	\caption{Illustration of the tokenization for $n=3$ and $m=2$. Starting from $o_1$, performing action $a_1$ will result in the next observation $o_2$ and the reward $r_2$. The constant separator tokens $t_5$ and $t_{12}$ are inserted to indicate the start of a new environment step.}
	\label{fig:tokenization}
\end{figure}

The distribution over a sequence of observations, actions, and rewards can be modeled by tokenizing the sequence first.
This is done by assigning a single integer (token) per scalar element (see \citet{reed2022generalist} for details), as illustrated in Fig.\ \ref{fig:tokenization}.
Thus, an $n$-dimensional observation is represented by a sequence of $n$ integers $(t_1,...,t_n)$, an $m$-dimensional action is represented by a sequence of $m$ integers $(t_1,...,t_m)$, and a reward is represented by a single integer.
While \citet{janner2021offline} follow a similar per-dimension tokenization scheme, \citet{chen2021decision} instead only use one token per state or action, obtained with a learned projection.

In the generalist experiments in sections \ref{sec:finetuning_generalist_results} and \ref{sec:zero-shot-generalization-results}, the TDM is used for predictions in multiple environments that have observation and action spaces of different dimensionalities.
All of these are translated into sequences of tokens (although of different per-timestep length depending on the dimensionality), which provide a unified interface to the TDM.

\subsection{Model Predictive Control (MPC)}
\label{sec:background_mpc}
MPC \citep{RICHALET1978413, garcia1989model, schwenzer2021review} refers to a group of control algorithms that make use of a model of the environment to choose the action in the current step.
Assume we have a model of the environment which at time $t$ allows us to predict the next $N$ observations $o^{(i)}_{t+1},...,o^{(i)}_{t+N}$ resulting from applying a sequence of $N$ actions $A^{(i)} = a^{(i)}_{t},...,a^{(i)}_{t+N-1}$.
This model allows us to predict a distribution
% \begin{align}
$
    P\left(o^{(i)}_{t+1},...,o^{(i)}_{t+N}|a^{(i)}_{t},...,a^{(i)}_{t+N-1}, o_t, h_t\right)
$
% \end{align}
of future observations, given the actions, a start observation $o_t$, and, depending on the model, a history $h_t$ of earlier observations, rewards, and actions, of arbitrary length.
We also call $N$ the planner horizon.
In its simplest form, given a set of candidate action sequences $\{A^{(1)},...,A^{(K)}\}$, an MPC controller compares these in terms of an objective function $f$, and chooses the first action of the action sequence that maximizes $f$:
\begin{align}
    a_t = a^{(k)}_t \ \text{,} \quad k = \argmax_i \mathbb{E} \left[f\left(o^{(i)}_{t+1},...,o^{(i)}_{t+N}, a^{(i)}_{t},...,a^{(i)}_{t+N-1}\right)\big| a^{(i)}_{t},...,a^{(i)}_{t+N-1}, o_t, h_t \right] \quad \text{.}
\end{align}
For the experiments in section \ref{sec:zero-shot-generalization-results}, the objective function $f$ explicitly depends on the rewards, but the rewards are not a deterministic function of the observations and actions.
In these cases, we use the TDM to predict a distribution
% \begin{align}
$
    P\left(r^{(i)}_{t},...,r^{(i)}_{t+N-1}, o^{(i)}_{t+1},...,o^{(i)}_{t+N}|a^{(i)}_{t},...,a^{(i)}_{t+N-1}, o_t, h_t\right)
$
% \end{align}
of both future observations and rewards, and then choose
\begin{align}
    a_t = a^{(k)}_t \ \text{,} \quad k = \argmax_i \mathbb{E} \left[f\left(r^{(i)}_{t},...,r^{(i)}_{t+N-1},o^{(i)}_{t+1},...,o^{(i)}_{t+N}, a^{(i)}_{t},...,a^{(i)}_{t+N-1}\right)\big| a^{(i)}_{t},...,a^{(i)}_{t+N-1}, o_t, h_t \right] \ \text{.}
\end{align}

\section{Method}
\label{sec:method}
At test time, the output of the transformer sequence model discussed in section \ref{sec:background_modelling_trajectory} is conditional on the sequence of tokens it has been prompted with.
This allows us to use it in different ways.
\begin{itemize}
    \item Condition on $(h_t, o_t)$, obtain $r_t$: Reward model
    \item Condition on $(h_t, o_t, r_t)$, obtain $a_t$: BC policy
    \item Condition on $(h_t, o_t, r_t, a_t)$, obtain $o_{t+1}$: Dynamics model (TDM)
\end{itemize}
The policy, reward model, or dynamics model are just different views on the same sequence model.
In the present work, we use the sequence model as a TDM, i.e., we focus on the last case.
We can test multiple candidate $a_t$, and query the TDM for its prediction of the effect.

In fully observable first-order Markov environments, $o_{t+1}$ only depends on $(o_t, a_t)$, making the history $h_t$ redundant for single-environment model learning (in practice, we found that including $h_t$ has a positive effect, but it is small, see section \ref{sec:context_window}).
However, a single time step $(o_t, a_t)$ usually does not contain enough information to infer the dynamics of the environment the data is from.
In the zero-shot generalist model learning case (section \ref{sec:training_setups}), the model also has to ``identify'' the dynamics of the target environment.
Therefore, a history $h_t$ of interactions with the environment is needed as a sample of the system dynamics in this case.

\subsection{MPC}
\label{sec:method_mpc}
Apart from a brief study of prediction errors in section \ref{sec:prediction_errors}, in this work we test the quality of the TDM's predictions by using it to create behavior in a simple MPC loop.
The TDM is used within the MPC loop to predict the outcome of action sequences $A^{(i)}$, given the current observation as start, and the history of the MPC agent's interaction with the environment since the beginning of the episode.

\paragraph{MPC with random shooting planner:}
\label{sec:mpc_with_random_shooting}
For most of the experiments, the candidate action sequences $A^{(i)}$ are independent of the observation, and randomly sampled from temporally correlated Brownian noise (see appendix \ref{sec:noise_discussion} for a discussion of this) with drift $0$ and variance $2$.
For the environments in this work, actions are clipped to the unit box $[-1,1]^m$, which is therefore symmetrically covered, with a slight bias for bang-bang control.

\paragraph{MPC with proposal:}
\label{sec:method_mpc_proposal}
For one experiment reported in section \ref{sec:humanoid_proposal}, we use a proposal policy $\pi(a|o)$ to obtain mean actions as a function of the observation predicted by the TDM, and then add temporally correlated Brownian noise with drift $0$ and varying levels of variance as shown in Fig.\ \ref{fig:expert_humanoid_with_proposal}.

\paragraph{Objective functions:}
For most environments used in this work, the reward can be obtained as a function $R(o')$ of predicted observations.
In these cases, we use the TDM to predict future observations, and then select actions to maximize the undiscounted future reward
\begin{align}
    f\left(o_{t+1},...,o_{t+N}, a_{t},...,a_{t+N-1}\right) = \sum_{k=1}^N R(o_{t+k})\quad \text{.}
\end{align}
For the \texttt{procedural walker} environments (see section \ref{sec:procedural_walker}), the reward can not be obtained from observations.
In these cases, we use the transformer sequence model to not only to predict future observations $o_t$, but also future rewards $r_t$.
We then use the objective function
\begin{align}
    f\left(r_{t},...,r_{t+N-1},o_{t+1},...,o_{t+N}, a_{t},...,a_{t+N-1}\right) = \sum_{k=1}^N r_{t+k}\quad \text{.}
\end{align}
We briefly discuss the performance difference between these two approaches in appendix \ref{sec:predicted_reward_comparison}.

\subsection{Training setups}
\label{sec:training_setups}
We consider multiple training setups that probe the model's ability to learn the dynamics for a single environment from experience in this environment, and also test its ability to generalize experience from previous environments to unseen environments.
This is described in the following.
\paragraph{Specialist model:}
For the experiments in section \ref{sec:expert_model_results}, we train the model with trajectories recorded in the same environment that we then use the model for MPC in.
We refer to this as the specialist model learning case.

\paragraph{Generalist model:} We consider two generalization scenarios.
\begin{itemize}
\item \textbf{Few-shot:}
For the experiments in section \ref{sec:finetuning_generalist_results}, we pre-train the model on a number of environments, and then fine-tune it on the unseen environment that we then use the model for MPC in.\\
\item \textbf{Zero-shot:}
For the experiments in section \ref{sec:zero-shot-generalization-results}, we train the model on a number of environments, and then use the model for MPC in an unseen environment without any fine-tuning.
\end{itemize}

\subsection{Environments}
\label{sec:environments}
\subsubsection{DeepMind control suite}
For the specialist experiments in section \ref{sec:expert_model_results} and the generalist fine-tuning experiments in section \ref{sec:finetuning_generalist_results}, we use control environments from the DeepMind control suite \citep{tassa2018deepmind}.
We use $3$ environments of increasing difficulty (\texttt{cartpole}, \texttt{walker}, \texttt{humanoid}) for the specialist experiments. For the generalist fine-tuning experiments, we pre-train on $28$ control suite environments, another $28$ versions of these environments with randomized parameters, and $24$ environments from the procedural walker universe (see below).

We can view these $80$ diverse control environments as samples of a high-dimensional space of environments.
$80$ samples are not nearly enough to densely cover this space.
Nevertheless, we are able to demonstrate a generalization effect for few-shot generalization to an unseen environment.

\subsubsection{The procedural walker universe of environments}
\label{sec:procedural_walker}
For the zero-shot generalization experiments in section \ref{sec:super_section_for_generalist_results} however, we need ``denser coverage'' of environments during training.
For this, we make use of the \texttt{procedural walker} universe of environments.
This setting was not purpose-created for the present work, but is unpublished so far.

\begin{figure}[h]
  \centering
  \begin{tabular}[c]{ccccc}
    \multirow{2}{*}[14pt]{
    \begin{subfigure}{0.5\textwidth}
      \includegraphics[width=\textwidth]{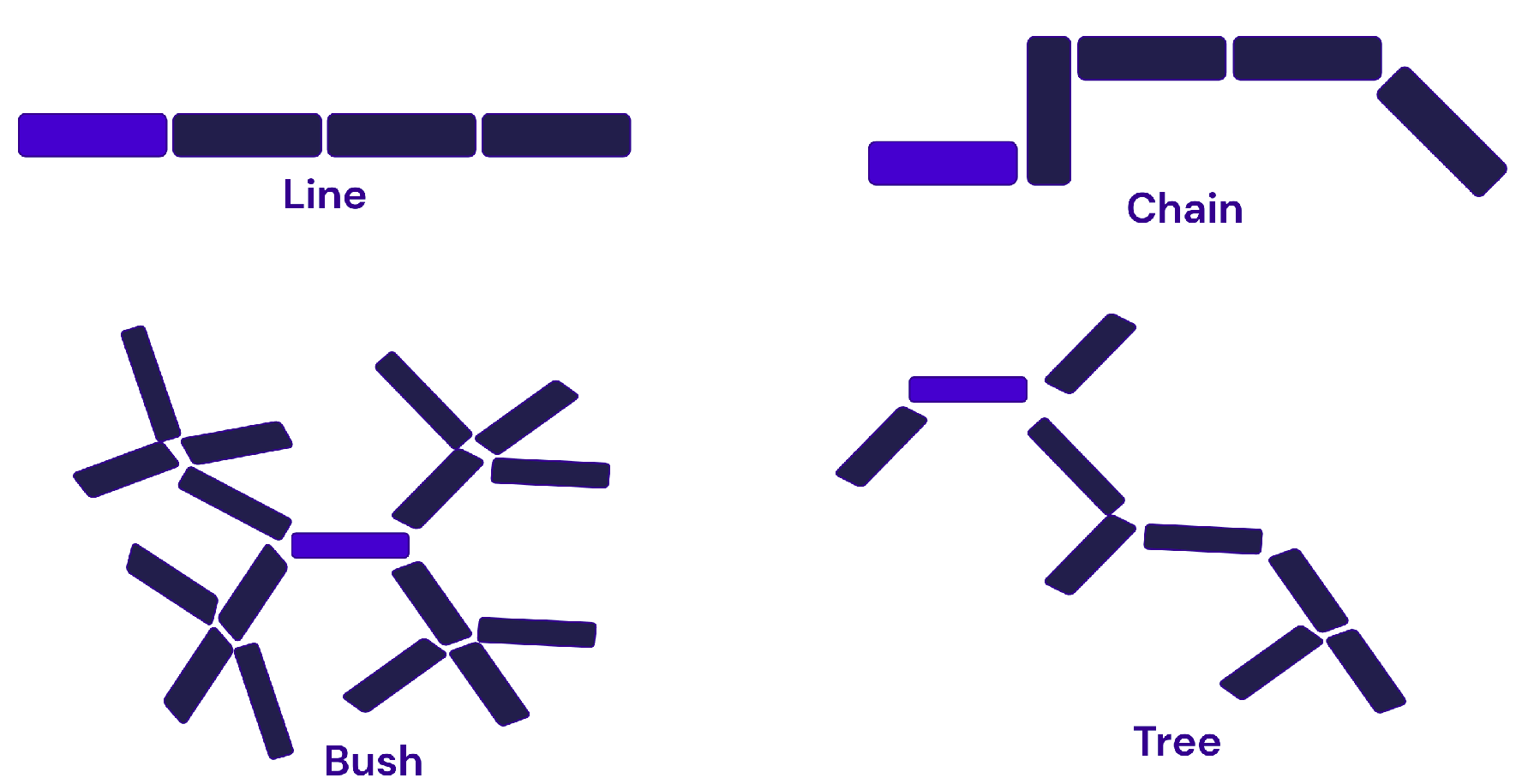}
    %   \caption{}
      \label{fig:proc_walker_overview}
    \end{subfigure}
}&
   \begin{subfigure}[c]{0.12\textwidth}
      \includegraphics[width=\textwidth]{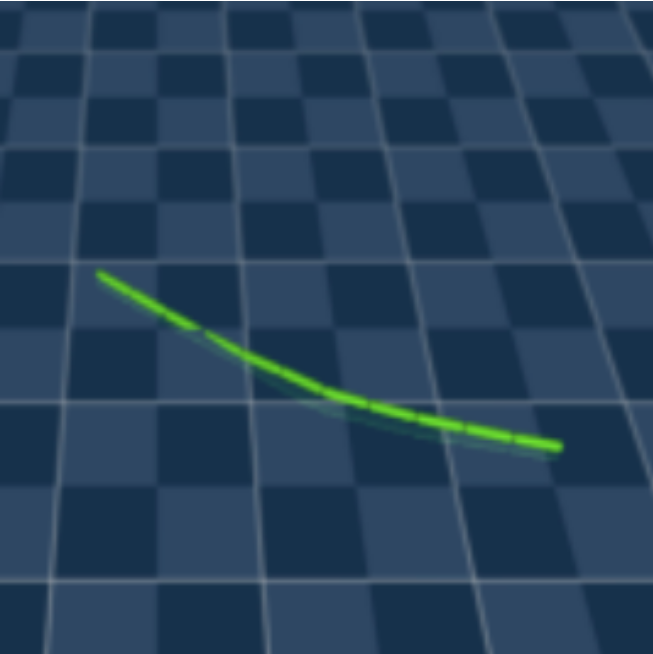}
      \caption{\texttt{Line}}
      \label{fig:proc_walker_line}
    \end{subfigure}&
    \begin{subfigure}[c]{0.12\textwidth}
      \includegraphics[width=\textwidth]{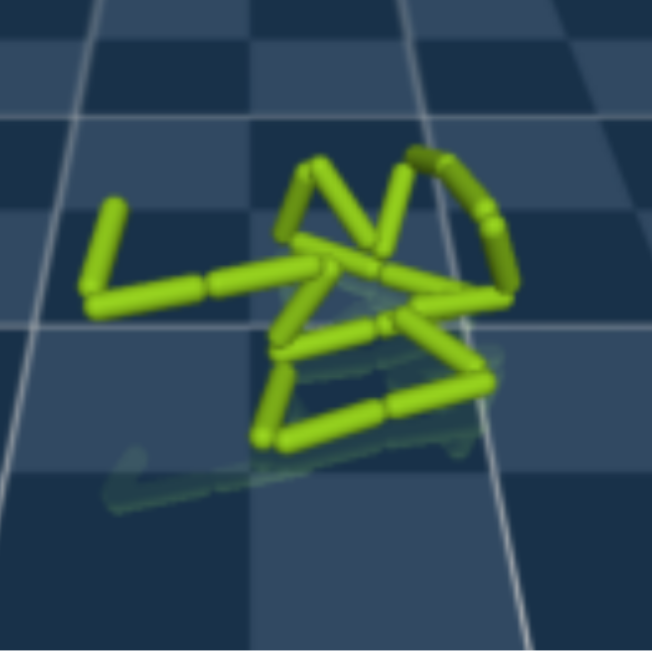}
      \caption{\texttt{Chain}}
      \label{fig:proc_walker_chain}
    \end{subfigure}\\
    &
        \begin{subfigure}[c]{0.12\textwidth}
      \includegraphics[width=\textwidth]{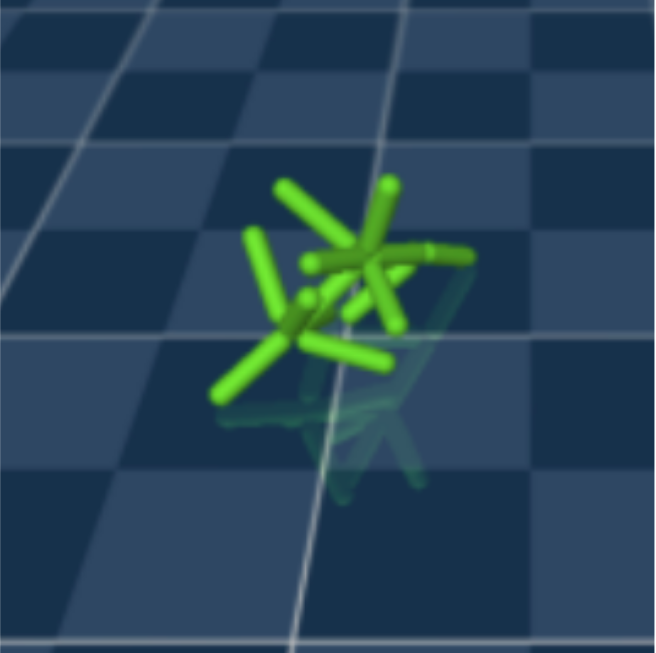}
      \caption{\texttt{Bush}}
      \label{fig:proc_walker_bush}
    \end{subfigure}&
    \begin{subfigure}[c]{0.12\textwidth}
      \includegraphics[width=\textwidth]{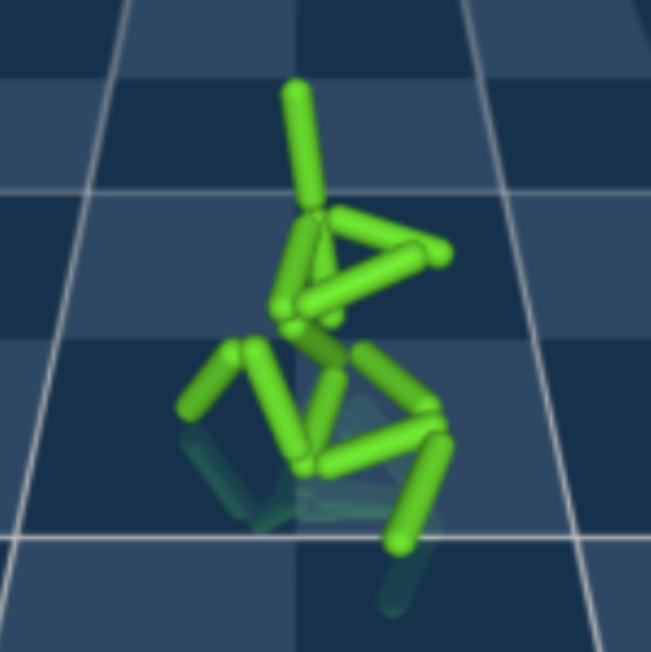}
      \caption{\texttt{Tree}}
      \label{fig:proc_walker_tree}
    \end{subfigure}\\
  \end{tabular}    
  \caption{The procedural walker universe.}
  \label{fig:procedural_walker_examples}
\end{figure}

The \texttt{procedural walker} universe contains procedurally generated locomotion environments with a diverse number of degrees of freedom (between $4$ and $20$ in our experiments) and diverse kinematic trees. The kinematic trees are constructed one link at a time.
The environments are divided into $4$ families.
Fig.\ \ref{fig:procedural_walker_examples} shows one example of each family.
For \texttt{line}, the next link is always added to the end of the previous limb, with the rotation axis being uniformly sampled from all possible rotation axes.
For \texttt{chain}, there are still either $1$ or $2$ links per limb, but one of $5$ attachment directions (6 axis-aligned directions minus the one pointing back into the limb) is randomly selected.
For \texttt{bush}, each limb has multiple limbs attached to it, with the only restriction that all links of a limb must be filled before moving on to one of its children.
Finally, for \texttt{tree}, this restriction is removed, and new limbs are randomly attached to any other limb in any direction, both selected uniformly at random.

The goal in all of these environments is to move in the positive $x$-direction with $\SI{1}{\meter/\second}$.
The exact reward is the one of the \texttt{run through corridor} example from \citet{tunyasuvunakool2020dm_control}.

\subsection{Training data}
\label{sec:training_data}
For all environments, the training data we use for model learning is collected by an expert or near-expert policy.
For the DeepMind control suite, we give a more detailed description of the resulting data distribution in section \ref{sec:data_dists}.
For our setting, this expert training data has, perhaps counterintuitively, a relatively challenging distribution.
As described in section \ref{sec:mpc_with_random_shooting}, the model is later queried with random action sequences following a distribution very different from the expert data it was trained on.

\section{Experiments}
\label{sec:experiments}
Fig.\ \ref{fig:experiments_overview} shows an overview of the experiments in this section.
The experiments demonstrate two aspects of TDMs.
\begin{enumerate}
\item Purple markers: TDMs are capable specialist control models, i.e., they are precise (compared to baselines) when trained with data from the target environment.
\item Yellow markers: TDMs are capable generalist control models, i.e., they show powerful few-shot or even zero-shot generalization capabilities.
\end{enumerate}
To this end, we show results in three different data regimes (specialist learning, generalist fine-tuning, generalist zero-shot).
These regimes are characterized by how much data from the target environment is available, and how much data from other environments is available.
Results are reported in sections \ref{sec:expert_model_results}, \ref{sec:finetuning_generalist_results},
and \ref{sec:zero-shot-generalization-results}, respectively.

We do report prediction errors of the TDM and baselines in section \ref{sec:prediction_errors}, but throughout this work mostly measure a model's quality by using it in a simple MPC loop with a random shooting planner, and measuring the reward of the resulting MPC agent.
This metric is tightly correlated with the model's usefulness for control (more so than, e.g., prediction accuracy).
We emphasize that, since the MPC algorithm is so simple, the resulting policy is often not state-of-the-art. We use MPC as a measuring tool for comparing model quality, not for building the best possible model-based agent.

\subsection{TDMs are capable single-environment models}
\label{sec:expert_model_results}
In the following, we evaluate the quality of TDMs when trained on sufficient data from the environment they are tested on.
We show that TDMs make accurate predictions that are suitable for planning for a range of difficult control tasks.
They consistently perform better than a number of baseline models in our experiments.
This finding remains robust if we switch to training data that was collected by an agent optimized for a different task (but in the same environment).
We also confirm these results by comparing prediction errors of the TDM to baselines in section \ref{sec:prediction_errors}.

\begin{figure}
    \captionsetup[subfigure]{justification=centering}
    \begin{subfigure}{.5\textwidth}
        \centering
        \includegraphics[width=\linewidth]{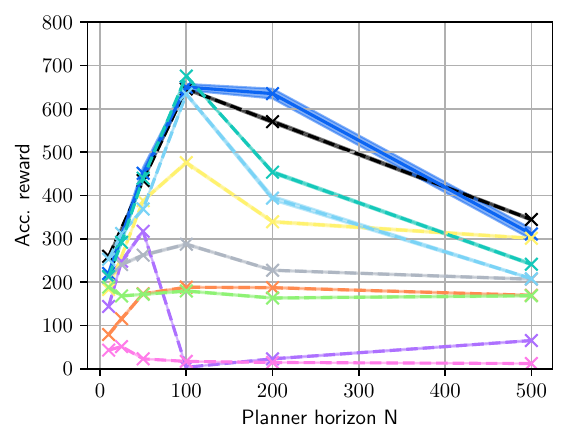}
        \caption{\texttt{cartpole swingup}}
        \label{fig:same_domain_results_cartpole}
    \end{subfigure}%
    \begin{subfigure}{.5\textwidth}
        \centering
        \includegraphics[width=\linewidth]{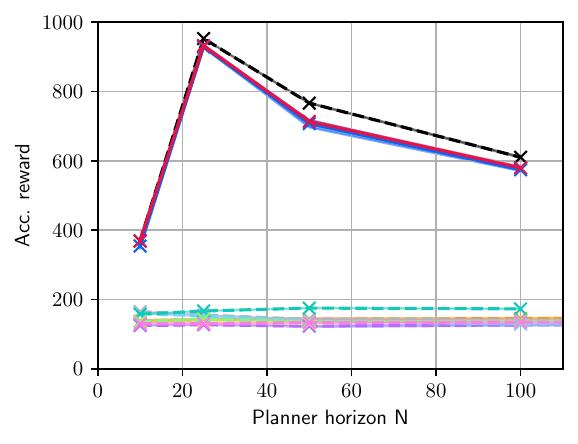}
        \caption{\texttt{walker stand}}
        \label{fig:same_domain_results_walker}
    \end{subfigure}
    \\
    \begin{subfigure}{.5\textwidth}
        \centering
        \includegraphics[width=\linewidth]{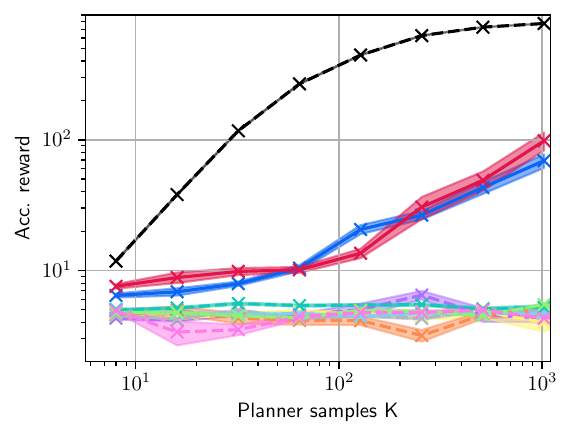}
        \caption{\texttt{humanoid stand}}
        \label{fig:same_domain_results_humanoid}
    \end{subfigure}
    \begin{subfigure}{.5\textwidth}
        \centering
        \vspace{-1.2cm}
        \includegraphics[width=0.55\linewidth]{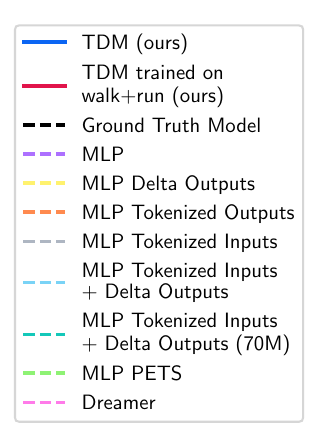}
        \label{fig:legend}
    \end{subfigure}
    \caption{Performance of TDMs and baseline models when trained on data from the environment they are tested on.
    We observe that TDMs consistently outperform baselines.
    This finding is robust when switching the training distribution to a different task in the same environment (red lines for walker and humanoid).
    We also compare with the ground truth models (black line).
    We evaluate the models by doing MPC with a very basic random shooting planner.
    The planner uses $K=128$ samples for cartpole, $K=64$ samples for walker, and horizon $N=20$ for humanoid.
    For very short planner horizons $N$, the planner is too myopic, and for very long horizons, the number of samples $K$ is insufficient for the random shooting planner to consistently discover a near-optimal action sequence.
    Therefore, when keeping $K$ fixed, there is an intermediate sweet-spot planner horizon.
    We report mean values averaged over at least $4$ episodes, shaded areas indicate $68\%$ confidence intervals.
    }
    \label{fig:same_domain_results}
\end{figure}
Fig.\ \ref{fig:same_domain_results} shows results for control tasks of increasing complexity from the DeepMind control suite \citep{tassa2018deepmind}: \texttt{cartpole swingup} (Fig.\ \ref{fig:same_domain_results_cartpole}), \texttt{walker stand} (Fig.\ \ref{fig:same_domain_results_walker}), and \texttt{humanoid stand} (Fig.\ \ref{fig:same_domain_results_humanoid}).
The data sets used contain $26762$, $18503$, and $12953$ episodes respectively, with $1000$ transitions each.
Since in this experiment, we want to test the model's ability to accurately fit the dynamics given sufficient data, we use large amounts of data to remove any data bottlenecks.
For more statistics on the data used, see section \ref{sec:data_dists}.
We also discuss prediction errors of the dynamics models in section \ref{sec:prediction_errors}.

We compare the TDM to the ground truth dynamics model, as well as different baseline dynamics models.
These baselines include a vanilla multilayer perceptron (MLP), MLPs that output the delta to the previous observation, MLPs with tokenized and embedded inputs, and MLPs with tokenized (categorical) outputs, as well as combinations thereof.
We also show results for a very large MLP with 70M parameters, a stochastic ensemble of MLPs (PETS, \citep{chua2018deep}), and the dynamics model of Dreamer V2 \citep{hafner2020mastering}.
Among the baselines, a combination of tokenized inputs and delta outputs (``MLP Tokenized Inputs + Delta Outputs'') seems to work best\footnote{We briefly zoom in on the relative performance of the MLP baselines using tokenized in- or outputs in section \ref{sec:tokenization_and_mlps}.}, and is on par with the TDM for shorter MPC planning horizons.
For longer planning horizons however, the TDM has an advantage.

For the more complex 6-DOF walker environment, the advantage of the TDM is even more pronounced: While the MPC agent based on the TDM reaches optimal performance, none of the baseline models is good enough to enable the MPC agent to reach better-than-random performance.
Finally, we find qualitatively similar results for the 21-DOF humanoid environment.
The TDM is the only model for which we observe non-random performance.

To rule out the possibility that the TDM (with ca.\ 70M parameters) outperforms these baselines (with ca.\ 400k parameters) simply because of its larger parameter size, for all environments we also include a version of the best-performing baseline that is much larger (70M parameters, ``MLP Tokenized Inputs + Delta Outputs (70M)'').
In all environments, the performance of this larger model is very similar to the performance of its smaller version, and again, the TDM has an advantage for longer planning horizons.

For cartpole and walker, the TDM performs on par with the expert ground-truth dynamics model.
For the $67$-dimensional humanoid, the TDM does not reach expert performance.
We use a random shooting planner for the experiments in Fig.\ \ref{fig:same_domain_results}, hence the TDM is queried with a state-action distribution that is very different from its training data (which comes from an expert policy, see appendix \ref{sec:data_dists}).
This distribution shift is challenging; while the TDM is still able to extrapolate perfectly in the cartpole and walker domains, we hypothesize that the extremely high dimensionality of humanoid makes it likely that the TDM is queried in areas of the state-action space that are simply not covered by its training data, making extrapolation almost impossible.
Having said that, the TDM is the only model with better-than-random performance in the humanoid domain, and we see a clear trend of increasing reward as we increase the number of samples $K$.
This indicates that the TDM's performance might increase further if the distribution shift is decreased.
We therefore discuss a planning approach using samples that are closer to the TDM's training distribution in the following.

\subsubsection{Including a proposal policy}
\label{sec:humanoid_proposal}
\begin{figure}[h]
	\centering
	\includegraphics[width=\columnwidth]{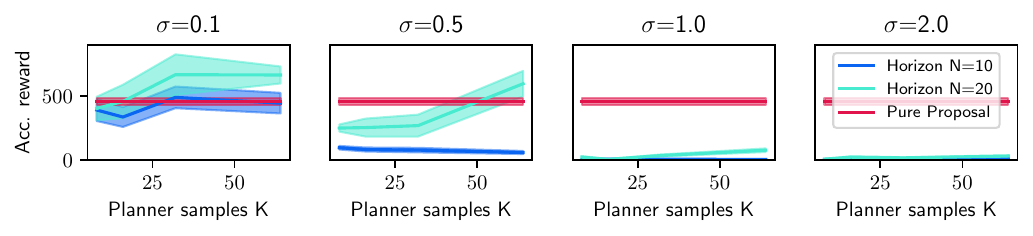}
	\caption{Using the TDM for MPC with a proposal policy for \texttt{humanoid stand}. The subfigures correspond to different levels of additive noise $\sigma$. Best results are obtained for moderate additive noise (this ensures that the bias of the proposal policy is not washed out) and larger horizons $N$ (this ensures that the planner does not become too myopic).
	The resulting MPC agent both works better than the pure proposal policy (red line), and needs less imaginary samples $K$ than the random shooting planner (see Fig.\ \ref{fig:same_domain_results_humanoid}).
    We report mean values averaged over at least $4$ episodes, shaded areas indicate $68\%$ confidence intervals.}
	\label{fig:expert_humanoid_with_proposal}
\end{figure}

We can make the planner use its budget of imaginary samples $K$ more efficiently by biasing the candidate action trajectories using a proposal policy, as described in section \ref{sec:method_mpc_proposal}.
Results for this are reported, for \texttt{humanoid stand}, in Fig.\ \ref{fig:expert_humanoid_with_proposal}.
As the proposal policy, we use the same transformer sequence model with the same weights that we also use as a TDM, but condition it as a BC policy at test time, as described in section \ref{sec:background_modelling_trajectory}.
The pure proposal policy is far from perfect, but useful as a bias.

Adding not-too-high amounts of additive noise to obtain candidate action sequences, and using a planning horizon $N$ that is not too myopic, the TDM can significantly improve on the proposal.
The model is able to consistently distinguish worse from better actions in the proposal-biased distribution; in fact the biased planner's performance approaches the asymptotic ($K\rightarrow\infty$) expert model's performance (Fig.\ \ref{fig:same_domain_results_humanoid}).
In this example, the hybrid approach of using the transformer sequence model both as a TDM and a BC policy outperforms each of these alone.

\subsubsection{Robustness against changes in training distribution}
\label{sec:changes_in_training_distribution}
As mentioned in section \ref{sec:training_data}, the distribution of the training data we use is strongly biased to expert performance, which, perhaps counterintuitively, is a challenging setup for learning a model that is then used for random shooting MPC.
For \texttt{walker stand} and \texttt{humanoid stand}, we also tested the TDM's performance after being trained on different distributions - namely expert data for \texttt{walker walk} and \texttt{run}, and \texttt{humanoid walk} and \texttt{run}, respectively.
As can be seen from Fig.\ \ref{fig:same_domain_results_walker} and Fig.\ \ref{fig:same_domain_results_humanoid}, the TDM's performance is largely unchanged by this.
This is additional evidence that TDMs are relatively robust against suboptimal training distributions.
Although not the focus of this work, to a certain extent this can also be seen as an example of the TDM generalizing across tasks (from \texttt{walk} and \texttt{run} to \texttt{stand}), but in the same environment.

The fact that our model outperforms the baselines considered in this chapter does not rule out the possibility that similar or even better performance is achievable with other architectures, including MLPs of different sizes and depths.
Our experiments show however that TDMs make accurate predictions that are suitable for planning for a range of difficult control tasks, in nontrivial learning settings that were very challenging for the baselines considered here.

\subsection{TDMs generalize to unseen environments}
\label{sec:super_section_for_generalist_results}
Next, we evaluate the quality of TDMs when trained on data from environments different from the one they are tested on.
We do this in two different settings:
We first show results of a generalist model that is pre-trained on a small number of unrelated control environments, and then fine-tuned on the unseen target environment (\texttt{cartpole}), in section \ref{sec:finetuning_generalist_results}.
We then report results for using a generalist model in zero-shot fashion in unseen environments in the \texttt{procedural walker} universe in section \ref{sec:zero-shot-generalization-results}.
For a discussion of our choice of environments, please refer to section \ref{sec:environments}.

\subsubsection{Few-shot generalization}
\label{sec:finetuning_generalist_results}
\begin{figure}
    \captionsetup[subfigure]{justification=centering}
    \begin{subfigure}{.5\textwidth}
        \centering
        \includegraphics[width=\linewidth]{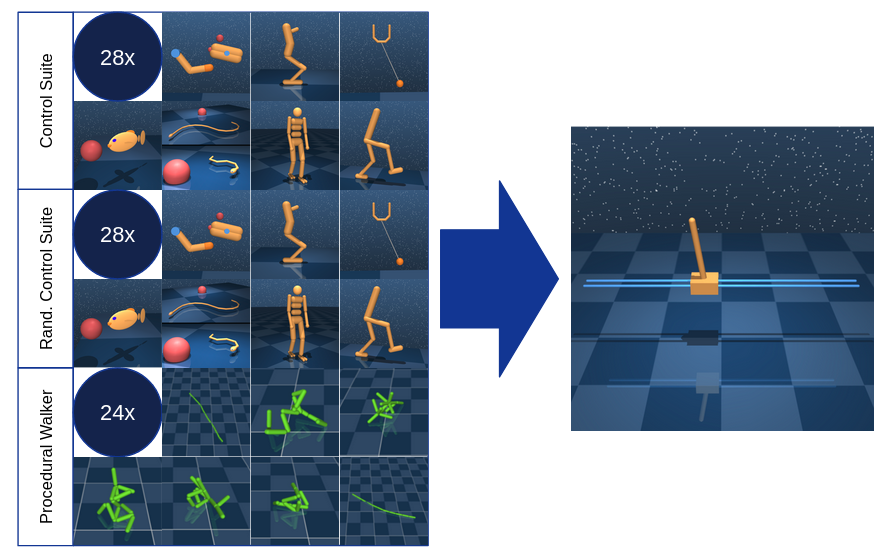}
        \caption{}
        \label{fig:cartpole_fewshot_exp_setup}
    \end{subfigure}%
    \begin{subfigure}{.5\textwidth}
        \centering
        \includegraphics[width=\linewidth]{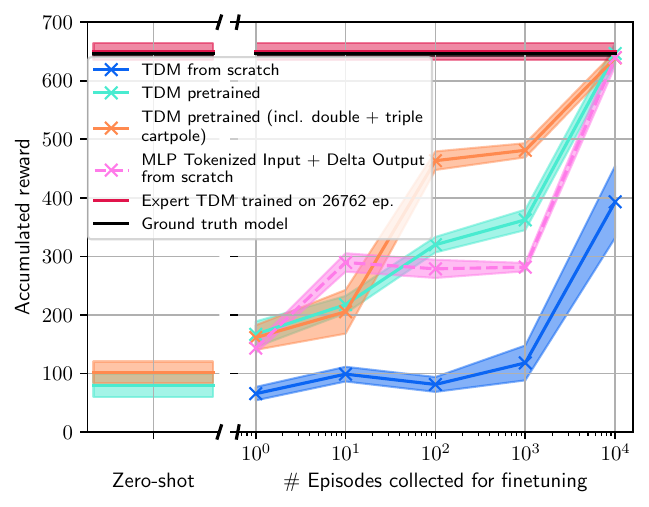}
        \caption{}
        \label{fig:cartpole_fewshot_results}
    \end{subfigure}
    \\
    \begin{subfigure}{1.0\textwidth}
        \centering
        \includegraphics[width=\linewidth]{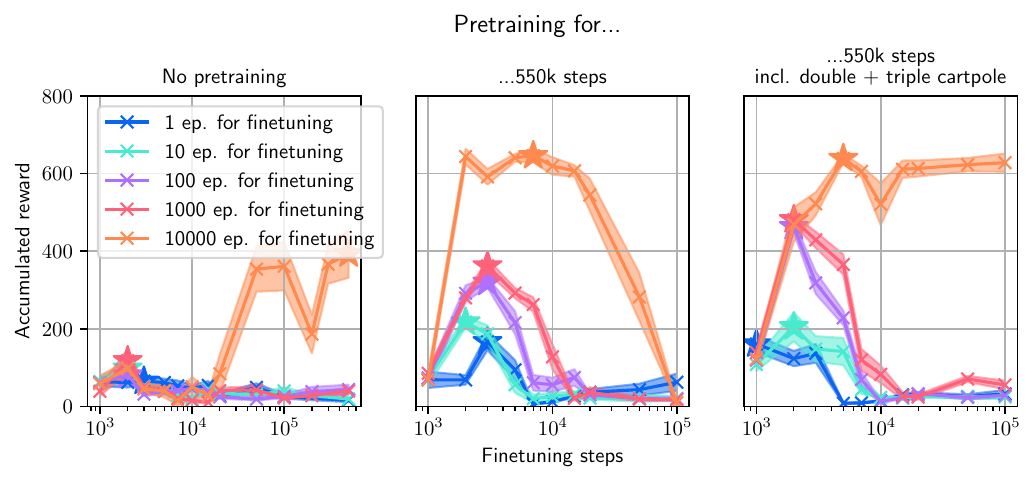}
        \caption{}
        \label{fig:cartpole_fewshot_finetuning_learning_curves}
    \end{subfigure}
    \caption{
    Few-shot generalization of TDMs.
    The results show that the TDM's generalization improves sample efficiency over low-expressivity baselines by almost $2$ orders of magnitude, and by $2$ to $3$ orders of magnitude over the from-scratch TDM.
    (a) We train a generalist model on ca. 100 environments that are unrelated to cartpole, fine-tune it with small amounts of data on cartpole, and test the resulting TDM on cartpole.
    (b) Model performances as a function of pre-training strategy and amount of data used for fine-tuning.
    There is a significant generalization effect, which further increases if we include double and triple cartpole data in our pre-training.
    Furthermore, in the medium data range, the fine-tuned TDM outperforms the best-performing MLP baseline.
    (c) Fine-tuning curves as a function of fine-tuning data and the pre-trained generalist model used. For each fine-tuning run, the best result is selected and shown in (b).
    Each episode contains $1000$ environment steps.
    The planner uses $K=128$ samples and horizon $N=100$.
    We report mean values averaged over $3$ independent fine-tuning runs and at least $4$ rollout episodes each, shaded areas indicate $68\%$ confidence intervals.
    }
    \label{fig:cartpole_generalization_results}
\end{figure}

We use a TDM as a generalist dynamics model.
The experimental setup is shown schematically in Fig.\ \ref{fig:cartpole_fewshot_exp_setup}.
We pre-train the model on $28$ environments from the DeepMind control suite, another $28$ randomized versions of the same environments, and $4\cdot6=24$ randomly created environments from the $4$ families of the \texttt{procedural walker} universe described in section \ref{sec:procedural_walker}.
None of these environments have any notable similarities with \texttt{cartpole}, our target environment.
We then fine-tune this model on different amounts of transition data from \texttt{cartpole}, and test the resulting model by using it for MPC with a simple random shooting planner, as in section \ref{sec:expert_model_results}.
This experiment is prototypical of a situation where the space of environments in which the model is supposed to generalize is only very sparsely covered by a relatively small number of pre-training environments.
Therefore, we unsurprisingly observe no zero-shot generalization, but we do observe significant few-shot generalization.

We vary the size $M$ of the fine-tuning data sets.
For each $M$, we fine-tune $3$ models on small data sets independently sampled from the full data set used in section \ref{sec:expert_model_results}.
For each $M$, we then optimize the number of fine-tuning steps independently.
These fine-tuning curves are shown in Fig.\ \ref{fig:cartpole_fewshot_finetuning_learning_curves}.
For each $M$, the average MPC performance is recorded after independently optimizing the number of training steps.
These optimized returns are indicated as stars in Fig.\ \ref{fig:cartpole_fewshot_finetuning_learning_curves}, and are shown as a function of $M$ in Fig.\ \ref{fig:cartpole_fewshot_results}.
The optimized returns reflect the TDM's performance as a function of the number of fine-tuning samples, rather than of the number of fine-tuning steps.
Comparing the performance of the fine-tuned generalist TDM to a TDM trained on the same small sets of data from scratch, we observe a significant few-shot generalization effect: We can obtain a similarly capable model with roughly $2$ to $3$ orders of magnitude less data.
As we increase the number of fine-tuning data, we approach the specialist model's (and ground truth's) performance reported in section \ref{sec:expert_model_results}.

\paragraph{Comparing to a different pre-training set.}
As mentioned earlier, the pre-training data set only contains data from environments that are entirely different from \texttt{cartpole}.
We also tested including data from \texttt{double cartpole} and \texttt{triple cartpole} in the pre-training data.
These environments are still quite different from \texttt{cartpole} (more degrees of freedom, different kinematics), but are arguably more related to cartpole than the environments originally in our pre-training set.
They can be considered to be closer to our target environment in the space of environments, potentially allowing the generalist model to few-shot-interpolate easier to the target environment.
Indeed, after including \texttt{double cartpole} and \texttt{triple cartpole} in the pre-training set, the results improve significantly over the original setting (see Fig.\ \ref{fig:cartpole_fewshot_results}).

\paragraph{Comparing to baselines - the data efficiency perspective.}
We also compare the generalization results with the best-performing MLP specialist from section \ref{sec:expert_model_results}.
This baseline is trained from scratch on the same data that was used for fine-tuning the TDM; again we use $3$ independent data sets and training runs each.
The MLP baseline (ca. 400k parameters, the TDM has ca. 77M) works better for very small amounts of data ($10$ episodes), but after that, the pre-trained generalist TDMs have a growing advantage.
In other words, given moderate amounts of data (ca. $100$ to $1000$ episodes), the fine-tuned generalist is the best model we were able to train in all of our experiments, including low-expressivity baselines.
In this regime, the generalist TDM needs almost $2$ orders of magnitude less data to achieve the same performance.
This is not because the generalist TDM has an inherently better sample efficiency (compare the from-scratch TDM to the MLP baseline), but rather it \textit{more than compensates} its initially lower sample efficiency by exploiting its capability to generalize from other environments.

The generalist TDM does not reach expert performance in the regime of $100$ to $1000$ episodes, but its much higher sample efficiency here has important practical utility, for example to warm-start exploration of a specialist agent.

\subsubsection{Zero-shot generalization}
\label{sec:zero-shot-generalization-results}
\begin{figure}[h]
	\centering
	\includegraphics[width=\columnwidth]{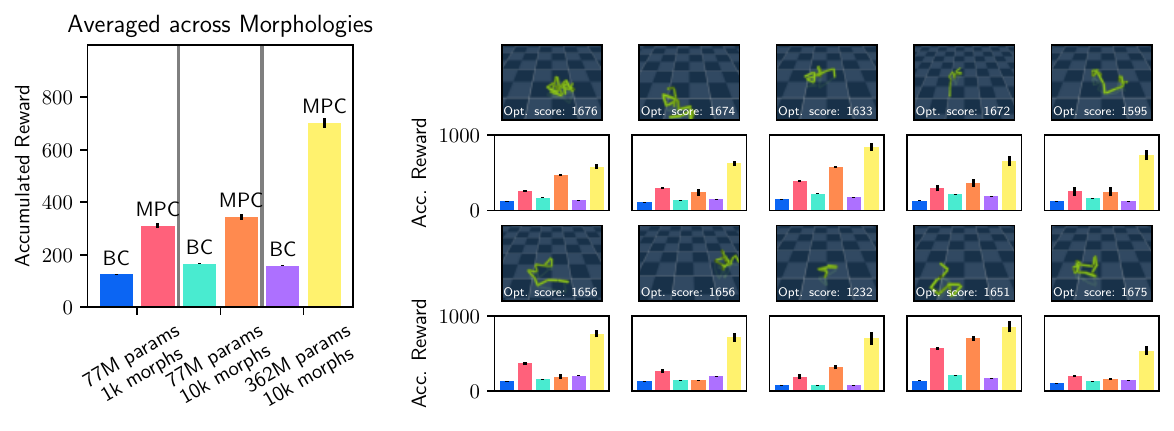}
	\caption{
    	Zero-shot generalization of TDMs.
    	The left side shows results averaged over all $10$ held-out test morphologies, the right side shows individual results.
    	We pre-train models of different size on expert data from either $1000$ or $10000$ different morphologies in the \texttt{chain} family of the \texttt{procedural walker} universe.
    	We then test the resulting generalist sequence model's performance as TDM in an MPC loop with a random shooting planner (as described in section \ref{sec:mpc_with_random_shooting}).
    	We compare this with using the same sequence model as a BC policy (see section \ref{sec:background_modelling_trajectory}).
    	We find that using the sequence model as TDM generalizes substantially better than using the same model as a BC policy.
    	This is especially true for the larger TDM with 362M parameters, which reaches roughly half of the maximum possible performance on average.
        Note that, at the start of each episode, we prompt the BC policy with a history of optimal behavior, providing it with privileged information.
        This is in contrast to the TDM, which we don't warm-start with any history.
    	We report mean values averaged over at least $4$ episodes, black bars indicate $68\%$ confidence intervals.
	}
	\label{fig:zeroshot_proc_walker}
\end{figure}

We use a TDM as a generalist model again, but now investigate its zero-shot generalization capabilities to an unseen environment.
As motivated in section \ref{sec:procedural_walker}, we use the  \texttt{procedural walker} universe for this, allowing for reasonable coverage of the space of environments the model is supposed to generalize in.
We train the model on either $1000$ or $10000$ randomly created morphologies from the \texttt{chain} family.
These morphologies are very diverse in their degrees of freedom (between $4$ and $20$) and kinematic trees (see section \ref{sec:procedural_walker}).
We then test the generalist TDM's performance on $10$ morphologies never seen during training.
The results are summarized in Fig.\ \ref{fig:zeroshot_proc_walker}.
The TDM zero-shot generalizes very well to unseen morphologies, especially for the larger model size tested.

In contrast to this, we measure no significant generalization effect when the same sequence model with the same weights is not used as TDM within an MPC loop with a random shooting planner, but as a BC policy.
While the TDM achieves roughly half of the optimal return, using the same model as BC policy does not generalize significantly.
If we train the same transformer model as a specialist BC policy on data from a single \texttt{procedural walker} environment, we achieve ca.\, $80\%$ of the optimal score on average.
This rules out insufficient model capacity or poor data quality as reason for the low performance of the BC policy in Fig.\ \ref{fig:zeroshot_proc_walker}; this can indeed be ascribed to weak generalization.
The transformer model is trained with expert data, providing high-quality data to the BC policy.
Additionally, at the start of each episode, we prompt the BC policy with a history of optimal behavior, providing it with privileged information.
In contrast, we don't warm-start the TDM with any history.

In this example, the TDM together with a planner that optimizes behavior generalizes better than the BC policy that directly models optimal behavior.
We speculate that there are at least two effects at play:
First, we observed that optimal behavior in the \texttt{procedural walker} universe can look very different depending on the morphology; while for some, a centipede-like walking motion is optimal, for others it is better to roll.
This means that identifying the dynamics from interaction (which is what the dynamics model (TDM) has to learn) might be an easier task than identifying optimal behavior, or at least continuing a prompt of optimal behavior, from interaction (which is what the behavior model (BC policy) has to learn).
Second, given an imperfect generalist sequence model, querying it repeatedly with random actions in an MPC loop might be more forgiving than directly querying it for actions.
The random actions create additional randomness in the behavior creating process that makes it less likely for the model to ``get stuck'' making wrong predictions.

Strong generalization is achieved in this experiment by using the generalist sequence model not (or not only) as policy, but as a TDM.
This across-environment generalization is in addition to the ``classic'' across-task generalization of dynamics models (see also section \ref{sec:changes_in_training_distribution}).

\section{Discussion}
\label{sec:discussion}
\noindent\textbf{Pixel observations}:
In this paper, we restrict our experiments to environments with state-based observations and did not consider pixel-based observations.
Apart from reducing need for computational resources, this was done in order to isolate generalization effects due to a transfer of a basic understanding of physics from generalization effects due to a transfer of perceptual capabilities.
That being said, pixel-based domains are an interesting and natural extension of our work for at least two reasons:
First, pixel-based observations open up our approach to more data sources, especially for real-world environments.
Second, images can contain richer context about the environment than states, allowing for faster system identification for generalization.
Fortunately, there are established techniques to tokenize image inputs for transformers, such as ViT \citep{dosovitskiy2020image} or VQGAN \citep{esser2021taming}.
Some of these approaches were already used with the Gato architecture we base this work on.
Furthermore, pixel-based domains require planning with predicted rewards, and initial experiments in appendix \ref{sec:predicted_reward_comparison} (and also the results in section \ref{sec:zero-shot-generalization-results}) indicate that our approach performs well in these cases.
We therefore believe that including pixel observations is a straightforward and natural extension of our work.

\noindent\textbf{Simple planner}:
As discussed earlier, the random shooting planner we used for MPC is a tool for comparing model quality.
As such, it is intentionally simple.
A planner optimized for performance likely could significantly improve the MPC reward.
We discussed one such example in Fig.\ \ref{fig:expert_humanoid_with_proposal}, where we used a proposal for planning.

\noindent\textbf{Training data}:
We use expert data (see also section \ref{sec:data_dists}) for training the dynamics models in this work.
As argued in section \ref{sec:training_data}, this is challenging for the models:
We train on expert data, but for random shooting MPC then query the models with state-action sequences distributed very differently.
In high-dimensional state-action spaces, this distribution shift makes it likely that the TDM will be queried in parts of the space for which it never ``saw'' any data.
Consistent with this, the TDM reaches expert level for cartpole and walker with random shooting, but not for the $67$-dimensional humanoid (Fig.\ \ref{fig:same_domain_results}).
After biasing the distribution with a proposal however, the TDM approaches expert performance on humanoid too (Fig. \ref{fig:expert_humanoid_with_proposal}).
This distribution shift could also explain why the Dreamer dynamics model performed well for policy improvement in \citet{hafner2019dream}, but not in Fig.\ \ref{fig:same_domain_results}.

\noindent\textbf{Utility of imperfect generalists}:
As is typical for generalization settings, the generalist TDM's predictions in the target environment are not perfect (see section \ref{sec:super_section_for_generalist_results}).
While imperfect, it is still significantly more informative than a non-generalizing model.
As such, it can be used as a bias to inform downstream learning algorithms, for example, to inform the exploration strategy of an RL agent in the target environment.

\noindent\textbf{Limits of generalization}:
Since the model has to interpolate in the space of environments in order to generalize, the pre-training data has to sample this space to some extent (section \ref{sec:zero-shot-generalization-results}).
For sparse sampling, fine-tuning might still be successful (section \ref{sec:finetuning_generalist_results}), but with sparse sampling and relatively complex targets, the generalization effect expectedly vanishes, as shown in section \ref{sec:walker_generalization_negative_example}.

\noindent\textbf{Model-free vs.\ model-based}:
This paper does not weigh in on whether model-based or model-free methods (or combinations, see section \ref{sec:humanoid_proposal}) are superior in every situation.
Indeed, model-free MPO reaches expert performance on cartpole after roughly $200$ episodes \citep{abdolmaleki2018maximum}, which is faster than our most efficient cartpole dynamics model, the fine-tuned TDM generalist, reaches expert model performance (see Fig.\ \ref{fig:cartpole_fewshot_results}).
Instead, we demonstrate that generalization is a powerful mechanism to speed up dynamics model learning (section \ref{sec:finetuning_generalist_results}), and we show that in some cases, model-based generalization does in fact outperform model-free
generalization (section \ref{sec:zero-shot-generalization-results}).

\noindent\textbf{Inference speed}:
Our current approach is limited by test-time inference speed.
The 77M model can predict roughly 500 tokens per second on a single Jellyfish TPU core, which, depending on the degree of parallelization, the dimensionality of the environment, the planner horizon $H$, and the number of planner samples $K$, can translate into environment step durations of tens of seconds in extreme cases.
Apart from increasing parallelization (down to one planner sample per core), this can likely be optimized significantly by using more sample-efficient planning algorithms than random shooting, an example of which was discussed in section \ref{sec:humanoid_proposal}.
The TDM itself can also be optimized for speed; a straightforward starting point is the context window size (inference time scales quadratically with window size), which can be reduced significantly without losing much of the performance, as shown in section \ref{sec:context_window}.
Finally, while the present work uses transformers as an example showing that generalist dynamics models exist at all, future research potentially will uncover completely different model architectures with similar generalization capabilities but faster inference speed.
On that note, we briefly discuss the possibility that tokenization could benefit non-transformer models in section \ref{sec:tokenization_and_mlps}.

Having said all that, fundamentally we propose to use transformer sequence models as large, expressive, generalist dynamics models that are not primarily optimized for speed.
A more principled way to resolve this trade-off between expressiveness and speed could be distillation:
Large general foundation models could be expressive and slow, but would then be distilled into light-weight specialists for specific tasks.
This could be done at several points along the execution pipeline: The TDM could be distilled into a dynamics model, or the MPC agent could be distilled into a policy.

\section{Conclusion}
\label{sec:conclusions}
We investigate using transformers as dynamics models (TDMs).
We demonstrate two aspects of TDMs in the experiments:
First, TDMs are generalist dynamics models, i.e., they generalize well to unseen environments, which we demonstrated both in the few-shot and in the zero-shot case.
Second, TDMs are capable specialist models, i.e., they are precise when learning from environment-specific data.

We believe that these properties make TDMs a promising ingredient for a foundation model of robotics and control.
As argued earlier, while we mostly focus on TDMs in this paper, using transformers as dynamics models or policies is not mutually exclusive.
A combination, like planning with proposals, might be the most efficient way to make use of the detailed and generalizable knowledge aggregated by a transformer that models the joint distribution of observations, actions, and rewards.

\section*{Acknowledgments}
We would like to thank Abbas Abdolmaleki, Philemon Brakel, Oliver Groth, Tuomas Haarnoja, Ben Moran, Francesco Nori, Scott Reed, and Dhruva Tirumala for insightful discussions and feedback.

% Bibliography components
\bibliographystyle{abbrvnat}
\bibliography{main}

\newpage
\appendix
\section*{Appendix}
\section{Training data distribution for DeepMind control suite}
\label{sec:data_dists}
\begin{figure}[h]
	\centering
	\includegraphics[width=\columnwidth]{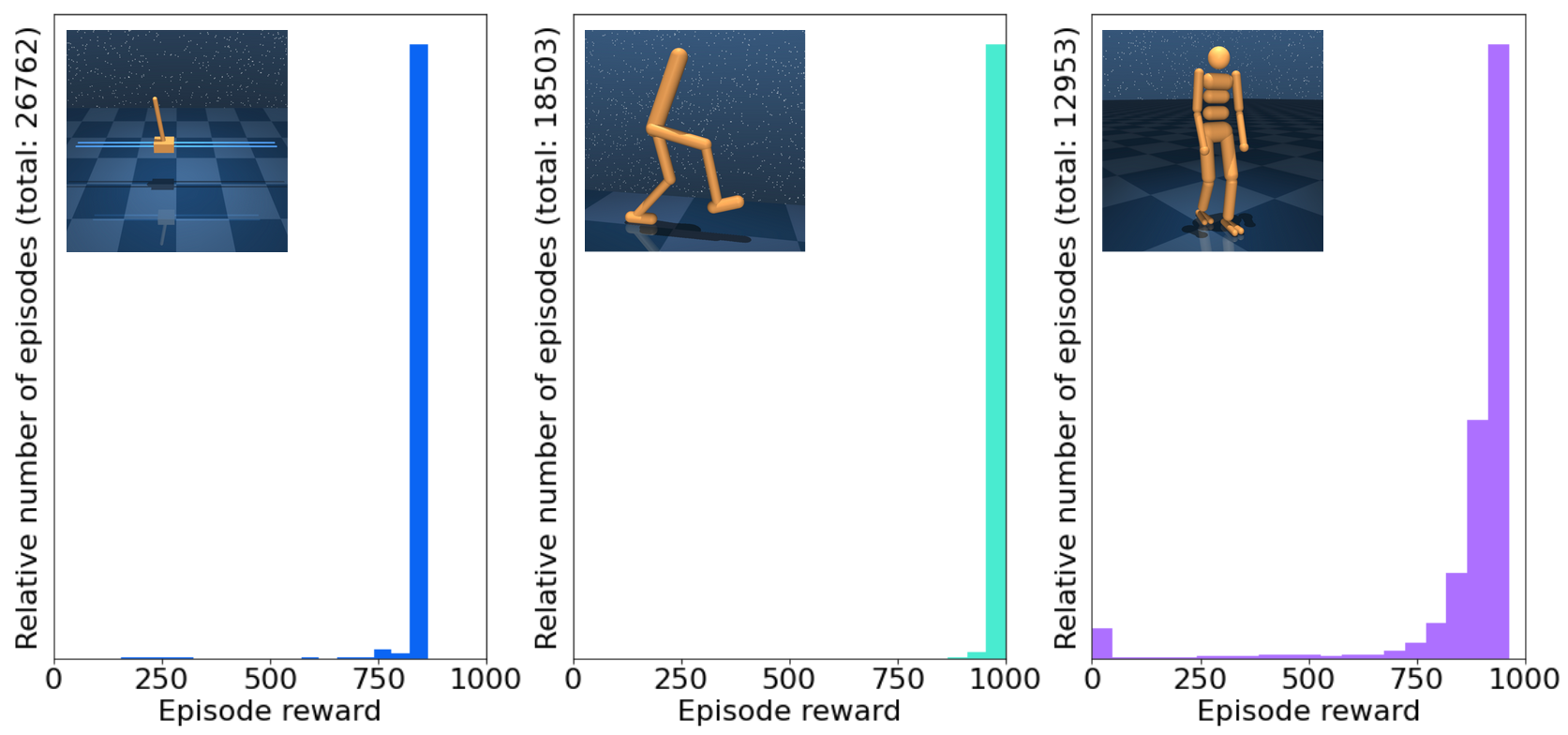}
	\caption{Distribution of episode rewards of the transition data used to train the models for \texttt{cartpole}, \texttt{walker}, and \texttt{humanoid} in section \ref{sec:expert_model_results}.}
	\label{fig:specialist_models_data_distribution}
\end{figure}

For the experiments with specialist models (section \ref{sec:expert_model_results}), Fig.\ \ref{fig:specialist_models_data_distribution} shows the distribution of episode rewards in the data used.
The data consists of mostly expert behavior.
As discussed in section \ref{sec:changes_in_training_distribution}, using mostly expert data poses a challenge for learning a model that can accurately predict rollouts with random actions, as required for our MPC agent.
The expert training data follows a very different distribution than the random actions at test time.

\section{Use of Brownian noise for random shooting MPC}
\label{sec:noise_discussion}
As described in section \ref{sec:mpc_with_random_shooting}, the candidate action sequences $A^{(i)}$ for MPC with random shooting are sampled from temporally correlated Brownian noise.
As an interesting note on this, \citet{eberhard2022pink} investigated time-correlated action noise for exploration in Deep RL with SAC \citep{haarnoja2018soft} and MPO \citep{abdolmaleki2018maximum} in the DeepMind control suite.
They find that pink noise (with a power spectral density proportional to $f^{-\beta}$, where $\beta=1$, which is between uncorrelated white ($\beta=0$) and Brownian ($\beta=2$) noise) worked best.
While we did not investigate the optimal value of $\beta$ in detail, we indeed found in preliminary experiments that MPC with uncorrelated noise ($\beta=0$) did not perform well.
This is perhaps unsurprising in retrospect: Non-correlated noise makes it exponentially unlikely to obtain control inputs that consistently favor one direction over extended periods of time, which is required for successful control in the DeepMind control suite.
In other words, time correlation is a general, but very beneficial prior when searching optimal action sequences for the DeepMind control suite.

\section{Varied context window}
\label{sec:context_window}
\begin{figure}[h]
	\centering
	\includegraphics[width=\columnwidth]{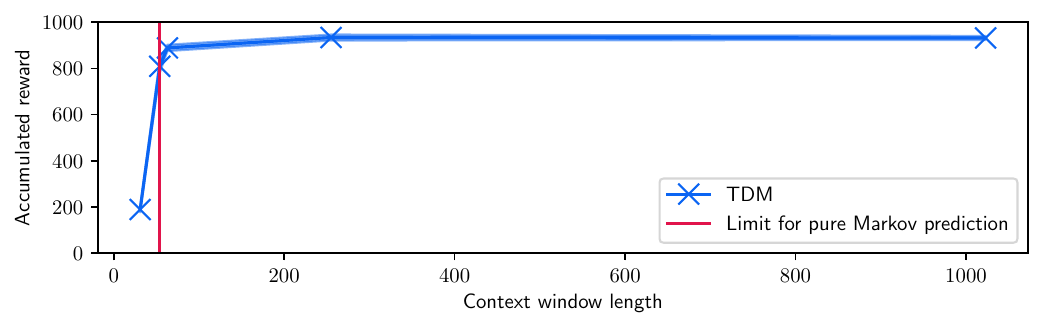}
	\caption{
	MPC performance of the specialist model for \texttt{walker stand} (see Fig.\ \ref{fig:same_domain_results_walker}) when using different context window sizes.
	The red line indicates the number of tokens that are needed to encode the previous observation and current action, i.e., the minimum context window needed in the strictly first-order Markov case.
	The results show that the model benefits from using additional context to some extent, but the difference is small compared to the difference to baseline models reported in Fig.\ \ref{fig:same_domain_results_walker}.
    The planner uses $K=64$ samples and horizon $N=25$.
    We report mean values averaged over at least $4$ episodes, shaded areas indicate $68\%$ confidence intervals.
	}
	\label{fig:context_window_experiment}
\end{figure}

The TDMs used throughout this work had a fixed context window length of $1023$ tokens.
For the \texttt{walker stand} task, Fig.\ \ref{fig:context_window_experiment} contains MPC rewards when using the TDM with varied context window length.
We observe that the performance is only very slightly affected by decreasing the context window size, until the window size becomes so small that it contains less than a single step.
This indicates that the TDM's performance does not predominantly rely on having a multi-step history as input.

\section{Performance of planning with predicted rewards}
\label{sec:predicted_reward_comparison}
As discussed at the end of section \ref{sec:method_mpc}, for most of the experiments in this work, the reward used for the objective function $f$ was computed from predicted future observations $o$ as $f\left(o_{t+1},...,o_{t+N}, a_{t},...,a_{t+N-1}\right) = \sum_{k=1}^N R(o_{t+k})$.
For the procedural walker experiments in section \ref{sec:zero-shot-generalization-results} however, the objective function $f\left(r_{t},...,r_{t+N-1},o_{t+1},...,o_{t+N}, a_{t},...,a_{t+N-1}\right) = \sum_{k=1}^N r_{t+k}$ is calculated using future rewards $r$ predicted by the TDM directly.

\begin{figure}
    \centering
    \includegraphics[width=0.5\linewidth]{"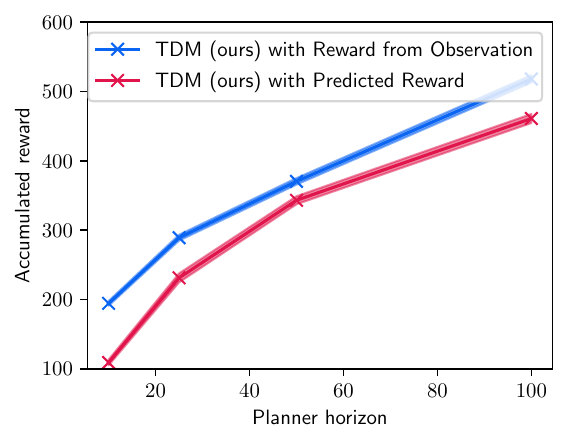"}
    \caption{Comparison of planning with rewards predicted by the TDM versus planning with rewards calculated from the observations predicted by the TDM for cartpole.
    The performance of directly planning with predicted rewards is only marginally worse.
    These results are for a smaller TDM architecture than the one we reported results for in Fig.\ \ref{fig:same_domain_results_cartpole}, hence there are small deviations from the values shown there.
    Shaded areas indicate $68\%$ confidence intervals.}
    \label{fig:predicted_reward_comparison}
\end{figure}

For cartpole swingup, Fig.\ \ref{fig:predicted_reward_comparison} compares these two approaches in terms of the reward of the MPC agent.
The performance of directly planning with predicted rewards is only marginally worse.
Note that these results were obtained with a smaller TDM architecture than the one we reported results for in Fig.\ \ref{fig:same_domain_results_cartpole}, hence there are small deviations from the values shown there.

As also discussed in section \ref{sec:discussion}, using the TDM in domains with pixel observations requires planning with predicted rewards.
This is because calculating the reward from pixel observations is usually impossible or at least infeasible.
The results in Fig.\ \ref{fig:predicted_reward_comparison} show that, at least for cartpole, switching to predicted rewards does not result in a large loss in performance.
Together with the points discussed in section \ref{sec:discussion}, this encourages us to hypothesize that an extension of TDMs to pixel-based domains might be straightforward.
We believe that an experimental investigation of this would be a natural avenue for future work.

\section{Example for unsuccessful generalization}
\label{sec:walker_generalization_negative_example}
\begin{figure}
    \captionsetup[subfigure]{justification=centering}
    \begin{subfigure}{.5\textwidth}
        \centering
        \includegraphics[width=\linewidth]{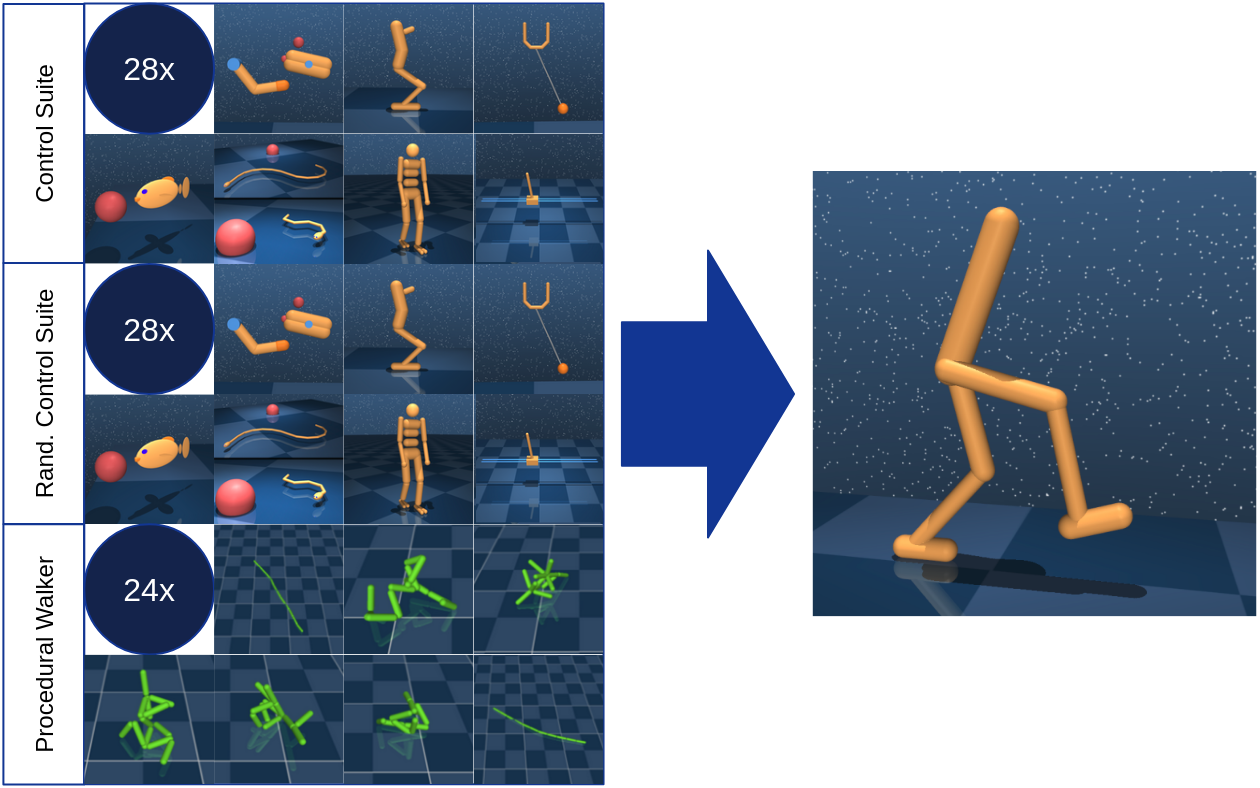}
        \caption{}
        \label{fig:walker_fewshot_exp_setup}
    \end{subfigure}%
    \begin{subfigure}{.5\textwidth}
        \centering
        \includegraphics[width=\linewidth]{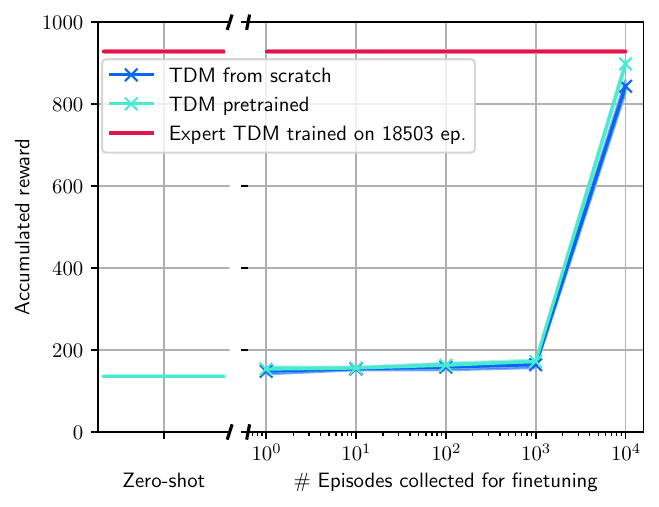}
        \caption{}
        \label{fig:walker_fewshot_results}
    \end{subfigure}
    \\
    \begin{subfigure}{1.0\textwidth}
        \centering
        \includegraphics[width=\linewidth]{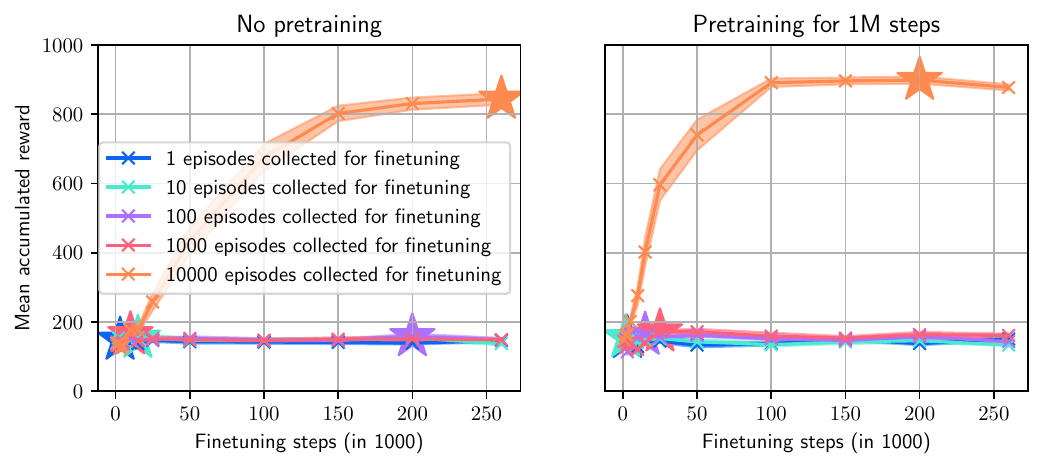}
        \caption{}
        \label{fig:walker_fewshot_finetuning_learning_curves}
    \end{subfigure}
    \caption{
    Example for unsuccessful few-shot generalization of TDMs.
    (a) We train a generalist model on ca. 100 environments that are unrelated to walker, fine-tune it with small amounts of data on walker, and then test the resulting TDM on walker.
    (b) Model performances as a function of pre-training strategy and amount of data used for fine-tuning.
    There is no significant generalization effect.
    (c) Fine-tuning curves as a function of fine-tuning data and the pre-trained generalist model used. For each fine-tuning run, the best result is selected and shown in (b).
    Each episode contains $1000$ environment steps.
    We report mean values averaged over $3$ independent fine-tuning runs and at least $4$ rollout episodes each, shaded areas indicate $68\%$ confidence intervals.
    }
    \label{fig:walker_failed_generalization_results}
\end{figure}

In Fig.\ \ref{fig:walker_failed_generalization_results}, we report an example where we did not observe a significant generalization effect.
Fig.\ \ref{fig:experiments_overview} puts this experiment in context with the other generalization experiments reported in sections \ref{sec:finetuning_generalist_results} and \ref{sec:zero-shot-generalization-results}.
We hypothesize that the combination of the sparse pre-training coverage of the space of environments ($80$ sample environments from the arguably huge space of environments that is covered by the DeepMind control suite), and the small amount of fine-tuning data (``small'' in relation to the relative complexity of the walker target environment), makes it impossible for the model to generalize.

Since cartpole is a simpler environment, the same amount of fine-tuning data provides a slightly better coverage of the target, and we observe a strong generalization effect with the very same pre-training coverage (section \ref{sec:finetuning_generalist_results}).
Note though that without using generalization effects, the problem would still be infeasible.

Finally, for the \texttt{procedural walker} results reported in section \ref{sec:zero-shot-generalization-results}, the target environments are arguably of similar complexity as the walker environment reported here, but the coverage of the space of environments in the pre-training data is better ($10000$ sample environments from the more structured space of environments that is covered by the \texttt{procedural walker chain} universe).

\section{Tokenization and MLPs}
\label{sec:tokenization_and_mlps}
\begin{figure}
    \centering
    \includegraphics[width=0.5\linewidth]{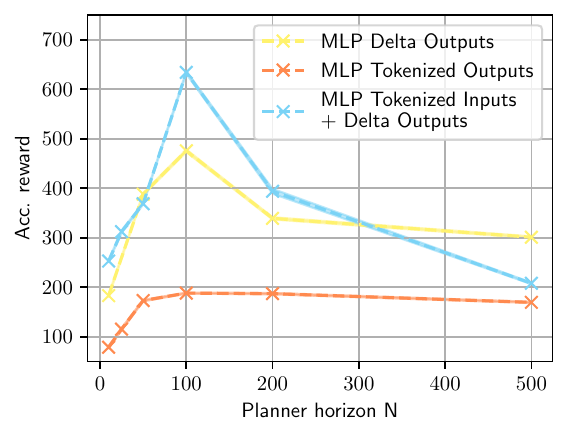}
    \caption{
    Using embedded tokens as inputs for an otherwise unchanged standard MLP increases its performance as a dynamics model.
    Changing the MLP to predict a categorical probability distribution over tokens decreases its performance.
    The planner uses $K=128$ samples for cartpole, $K=64$ samples for walker, and horizon $N=20$ for humanoid.
    We report mean values averaged over at least $4$ episodes, shaded areas indicate $68\%$ confidence intervals.
    }
    \label{fig:tokenization_ablation}
\end{figure}

We zoom in on some of the results for the MLP baselines reported in Fig.\ \ref{fig:same_domain_results_cartpole} in order to discuss the effect of tokenization.
Fig.\ \ref{fig:tokenization_ablation} shows the effect of adding tokenization of inputs or outputs of an otherwise unchanged MLP.
We find that using embedded tokens as inputs for an otherwise unchanged standard MLP increases its performance as a dynamics model, while changing the MLP to predict a categorical probability distribution over tokens decreases its performance.
While it is not the purpose of the present work to identify successful design choices that might translate from transformers to other architectures as well, using tokenized input is one example of this that works well in the present case.
This might be an interesting starting point for future investigations.

\section{Prediction errors}
\label{sec:prediction_errors}
\begin{figure}
    \centering
    \includegraphics[width=\linewidth]{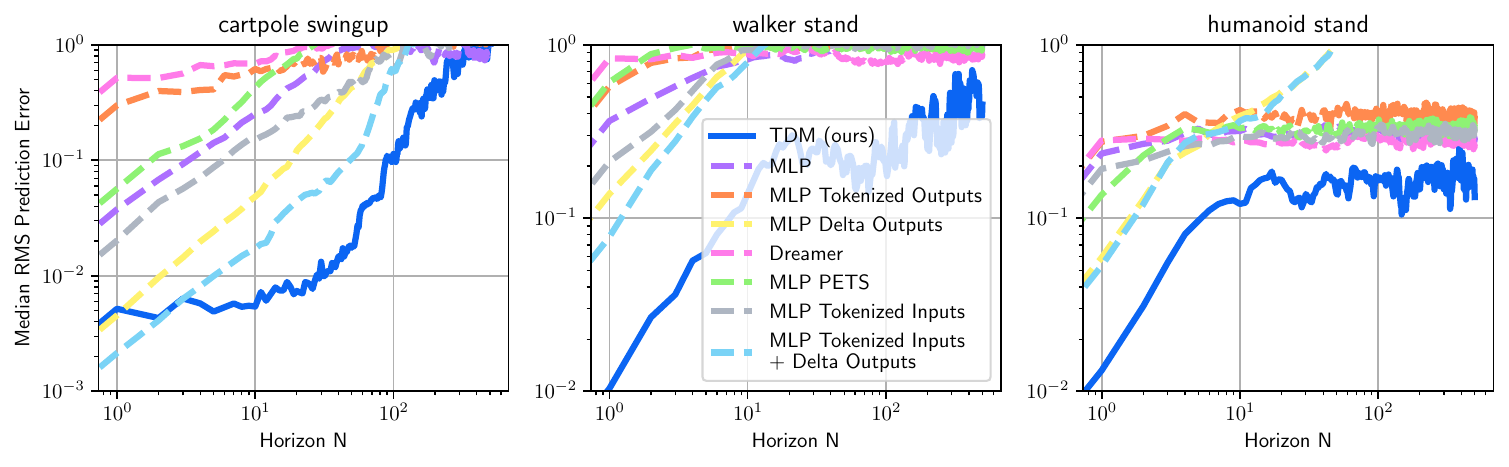}
    \caption{
    Prediction accuracies (RMS error leaving out velocities) for the models used in Fig.\ \ref{fig:same_domain_results}.
    Across all environments, the TDM is significantly more accurate than the baselines, especially for longer horizons $N$.
    The lines show median values taken over $30$ runs.
    Runs are collected by randomizing the initial state, and then executing random actions from the same distribution that the random shooting planner uses as well.
    For \texttt{walker stand} and \texttt{humanoid stand}, the baseline's prediction accuracy for horizons that are sufficient for effective planning is too low to accurately distinguish good from bad action sequences, resulting in the poor MPC performance observed in the results shown in Fig.\ \ref{fig:same_domain_results}.
    }
    \label{fig:prediction_errors}
\end{figure}

For the single-environment models, we reported the performance of the resulting MPC agent in Fig.\ \ref{fig:same_domain_results}.
Additionally, Fig.\ \ref{fig:prediction_errors} shows prediction errors for the models used in Fig.\ \ref{fig:same_domain_results}.
Across all environments, the TDM is significantly more accurate than the baselines.
This difference is especially pronounced for longer horizons $N$.
For the more complex environments \texttt{walker stand} and \texttt{humanoid stand}, the baseline's prediction accuracy for horizons that would be sufficient for effective planning ($N>20$) is too low to accurately distinguish good from bad action sequences, resulting in the poor MPC performance observed in the results shown in Fig.\ \ref{fig:same_domain_results}.

\end{document}